\documentclass[sigconf]{acmart}

\usepackage[ruled, linesnumbered]{algorithm2e} 
\usepackage{bm}
\usepackage{adjustbox}
\usepackage{color}
\usepackage{balance} 
\def\argmax{\mathop{\rm arg\,max}\limits}
\def\argmin{\mathop{\rm arg\,min}\limits}

\AtBeginDocument{%
  \providecommand\BibTeX{{%
    \normalfont B\kern-0.5em{\scshape i\kern-0.25em b}\kern-0.8em\TeX}}}

\copyrightyear{2021}
\acmYear{2021}
\setcopyright{acmcopyright}
\acmConference[MM '21] {Proceedings of the 29th ACM Int'l Conference on Multimedia}{October 20--24, 2021}{Virtual Event, China.}
\acmBooktitle{Proceedings of the 29th ACM Int'l Conference on Multimedia (MM '21), Oct. 20--24, 2021, Virtual Event, China}
\acmPrice{15.00}
\acmISBN{978-1-4503-8651-7/21/10}
\acmDOI{10.1145/3474085.3475614}

\begin{document}
\fancyhead{} 
\title{Delving into Deep Image Prior for Adversarial Defense: \\
A Novel Reconstruction-based Defense Framework}  

\author{Li Ding$^{1,2\ast}$, \ Yongwei Wang$^{2\ast}$, \ Xin Ding$^{2}$, Kaiwen Yuan$^{2}$ \\ \ Ping Wang$^{1}\dagger$,  \ Hua Huang$^{3}\dagger$, \ Z. Jane Wang$^{2}$}
\affiliation{\institution{$^{1}$ School of Information and Communications Engineering, Xi'an Jiaotong University \city{Xi'an} \country{China}}}
\affiliation{\institution{$^{2}$ Department of Electrical and Computer Engineering, University of British Columbia \city{Vancouver} \country{Canada}}  } 
\affiliation{\institution{$^{3}$ School of Artificial Intelligence, Beijing Normal University \city{Beijing} \country{China}}}
\affiliation{\institution{\{dinglijay, yongweiw, kaiwen, zjanew\}@ece.ubc.ca, xin.ding@stat.ubc.ca, ping.fu@xjtu.edu.cn, huahuang@bnu.edu.cn \country{}}
}
	
\thanks{$^\ast$Equal contribution, alphabetical order.\\
$^\dagger$Corresponding author.
}

\renewcommand{\shortauthors}{Ding and Wang, et al.}

\begin{abstract}
Deep learning based image classification models are shown vulnerable to adversarial attacks by injecting deliberately crafted noises to clean images. To defend against adversarial attacks in a training-free and attack-agnostic manner, this work proposes a novel and effective reconstruction-based defense framework by delving into deep image prior (DIP). Fundamentally different from existing reconstruction-based defenses, the proposed method analyzes and explicitly incorporates the model decision process into our defense. Given an adversarial image, firstly we map its reconstructed images during DIP optimization to the model decision space, where cross-boundary images can be detected and on-boundary images can be further localized. Then, adversarial noise is purified by perturbing on-boundary images along the reverse direction to the adversarial image. Finally, on-manifold images are stitched to construct an image that can be correctly predicted by the victim classifier. Extensive experiments demonstrate that the proposed method outperforms existing state-of-the-art reconstruction-based methods both in defending white-box attacks and defense-aware attacks. Moreover, the proposed method can maintain a high visual quality during adversarial image reconstruction.
\end{abstract}



\maketitle

\section{Introduction}
Recent studies show that deep convolutional neural network (DCNN) based classifiers are vulnerable to \textit{adversarial examples} \cite{szegedy14,FGSM,moosavi2016deepfool,MIFGSM,madry2018towards,transfer19,wang2021perception,zhang2020geometry}. By injecting small and imperceptible adversarial perturbations into natural images, such deliberately crafted adversarial examples can significantly degrade the performance of victim classifiers. For safety-critical applications, it is crucial to study adversarial defense methods to protect victim classifiers from being attacked by adversarial examples. Ideally, an effective defense method includes two desirable properties: 1) \textbf{attack-agnostic}, i.e., it can defend against different types of (even unknown) attacks. 2) \textbf{training-free}, i.e., the defense does not require to retrain the defense model even tested on a different dataset.         

Many defense methods have been proposed to counter adversarial attacks \cite{FGSM,madry2018towards,xiao2020one,xie2017mitigating,guo2017countering,mustafa2019image,xie2019feature,jia2019comdefend,naseer2020self,sutanto2020adversarial,dai2020dipdefend}. We can broadly categorize existing defense mechanisms into two classes: robustness-based defenses and reconstruction-based defenses. Robustness-based defenses attempt to robustify DCNN classifiers by retraining them on an augmented dataset, where adversarial examples are used as data augmentation \cite{FGSM,madry2018towards,xie2019feature,xiao2020one}. Such defenses display robustness to some specific attacks from which adversarial examples are generated. Unfortunately, these defenses are often computationally expensive to be practically applicable, particularly for large-scale datasets with high-dimensional images (e.g., ImageNet \cite{deng2009imagenet}). Besides, robustness-based defenses display vulnerability to attacks with larger perturbations. Moreover, the retrained classifiers generally have degraded performance in recognizing clean images \cite{FGSM,xie2019feature}.        

Compared to robustness-based defenses, reconstruction-based defenses are more scalable and thus more promising in practical applications \cite{xie2017mitigating,guo2017countering,jia2019comdefend,mustafa2019image,naseer2020self,sutanto2020adversarial,dai2020dipdefend}. Given an adversarial example, reconstruction-based defenses endeavor to recover a clean image that can be correctly recognized by the victim classifier. A majority of works require to train an auxiliary network to purify adversarial noise, (e.g., \cite{jia2019comdefend,naseer2020self}). However, such methods may overfit to the training dataset, which often hardly generalize to unseen datasets; Or they may fail even in the presence of data distribution shift. Moreover, these learning-based defenses usually suffer from certain attacks in the attack-agnostic setting. 

\begin{figure}[!htbp]
  \centering
  \includegraphics[width=0.6\linewidth]{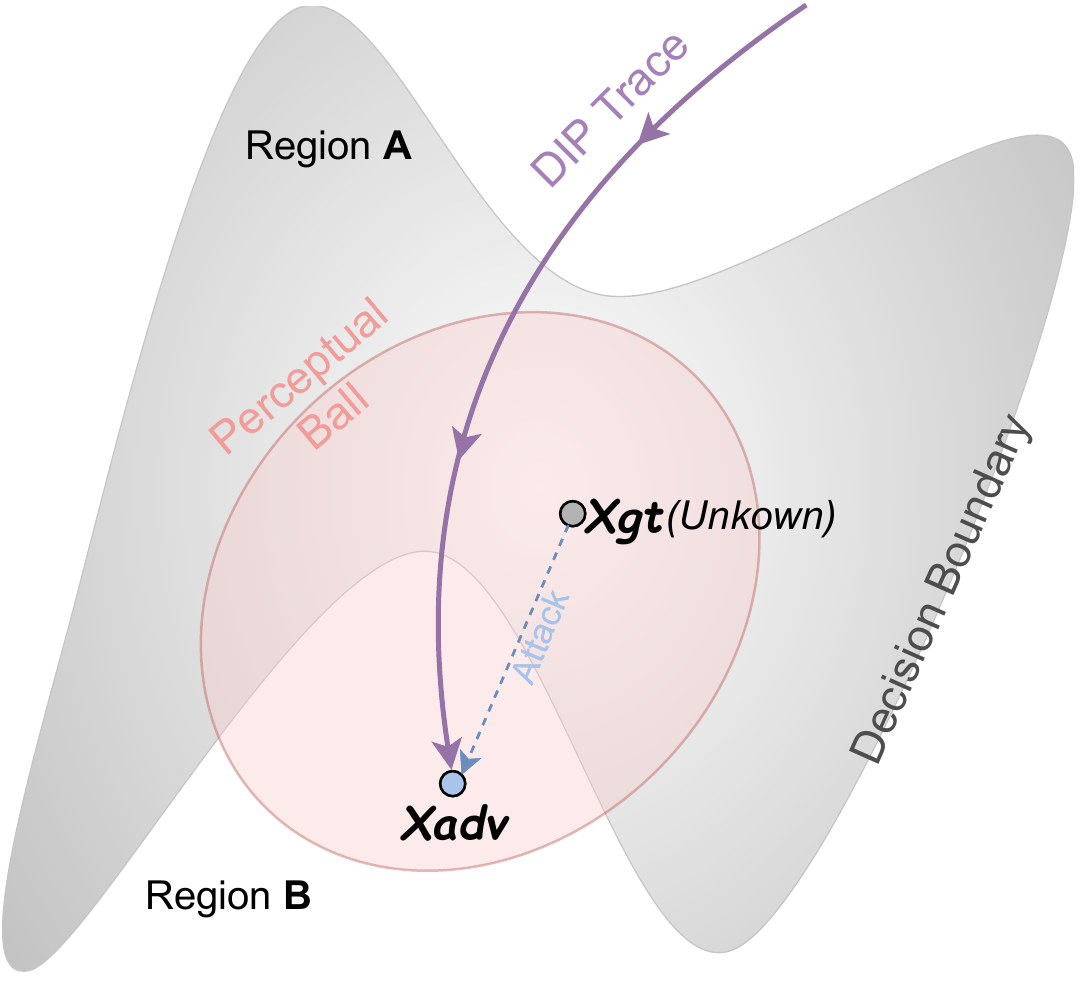} 
  \caption{Illustration of an image reconstruction trace from a DIP network specific to an adversarial example $\boldsymbol{x}_{adv}$. Starting from a random point in the image space, the reconstructed image gradually approaches the manifold of natural images (i.e., region A) and then reaches the manifold of adversarial images (i.e., region B).  Adversarial example $\boldsymbol{x}_{adv}$ is generated from a (unknown) clean image $\boldsymbol{x}_{gt}$ by injecting adversarial perturbations within a perceptual ball.  
  } 
  \label{fig:manifold_view}
\end{figure}

Therefore, more recent works try to explore the training-free reconstruction-based defenses \cite{kattamis2019exploring,shi2020unsupervised,dai2020dipdefend}. Instead of relying on an external prior learned from a dataset, existing training-free defenses exploit the internal prior (a.k.a. deep image prior (DIP) \cite{ulyanov2018deep}) from a single natural image. The state-of-the-art method is DIPDefend \cite{dai2020dipdefend}. It adopts a DIP generator (i.e., an untrained DCNN network) to reconstruct a clean image $\boldsymbol{x}$ from an adversarial example $\boldsymbol{x}_{adv}$. As shown in Fig.~\ref{fig:manifold_view}, given an $\boldsymbol{x}_{adv}$, a DIP network (specific to $\boldsymbol{x}_{adv}$) generates a sequence of reconstructed images which form the DIP trace. DIPDefend hypothesizes that the DIP trace passes through the natural image manifold (i.e., region A). To prevent the DIP network from overfitting to $\boldsymbol{x}_{adv}$ (i.e., entering region B), DIPDefend proposes a criterion to select one image (along the DIP trace) as its final reconstruction. However, since the criterion takes place in the image space, it cannot reliably alleviate the overfitting problem especially for adversarial examples that locate near the decision boundary (e.g., $\boldsymbol{x}_{adv}$ from C\&W attack \cite{carlini2017adversarial}).   
             
To effectively address the overfitting problem in DIP-based defenses, we propose to incorporate classifier decisions into the defense. By projecting DIP reconstructed images to the decision space, we can reliably localize the decision boundary such that we may reject images that are within the manifold of adversarial examples (e.g., region B in Fig.~\ref{fig:manifold_view}). Even with knowledge of the decision boundary, unfortunately, often it is infeasible to determine which images around the decision boundary should be used for final reconstruction. This is because these images may belong to different classes from the correct one. Instead, we propose to employ on-boundary images which can be constructed from cross-boundary images through linear search. We then propose a novel denoising strategy to process on-boundary images and obtain a reconstructed image that can be correctly predicted by the victim classifier. Meanwhile, the reconstructed image preserves a high visual quality. Specifically, we move on-boundary images a small step along the reverse direction to the adversarial example and obtain on-manifold images which are adversarial noise-free. These images contain reconstruction errors due to imperfect reconstruction of the DIP network. By stitching such on-manifold images together, we can largely suppress the reconstruction error and yield a clean image. 

To summarize, our main contributions are threefold:  
\begin{itemize}
	\item We propose a novel reconstruction-based adversarial defense method by delving into deep image prior. Our method is attack-agnostic and training-free. Specially, different from existing DIP-based defenses, we address the overfitting problem with a conceptually simple yet effective strategy. 
	
	\item The proposed method explicitly integrates the decision of the victim classifier into our defense. In the decision space, our method detects boundary images and reconstructs a clean image utilizing the proposed novel denoising method.

	\item We conduct extensive experiments on CIFAR-10 and ImageNet to demonstrate the defensive effectiveness of the proposed method. Experimental results show that our method achieves state-of-the-art defense performance both in defending against white-box attacks and defense-aware attacks. Moreover, the proposed method can also preserve high image visual quality. 
	
\end{itemize}

\section{Related Work}
In this section, we first introduce the adversarial attack problem. We then review related works on some popular adversarial attack/defense methods. 

Adversarial attacks aim to fool deep neural networks by generating adversarial examples \cite{szegedy14,FGSM}. Assume a DCNN classifier $f: \mathcal{D} \subseteq \mathbb{R}^d \mapsto \mathbb{R}^C$, where $\mathcal{D} =[0, 255]^d$. Given a clean image $\boldsymbol{x} \in \mathbb{R}^d$, the classifier correctly predicts its true class as $y \in \mathcal{Y}$, i.e., $y=\argmax_{c=1, 2, \cdots, \; C} f_c(\boldsymbol{x})$. Typically, an attacker performing non-targeted adversarial attacks attempts to find an adversarial example $\boldsymbol{x}_{adv}$ that can fool the classifier into making a wrong decision, 
\begin{equation}
\begin{split}
    \argmax_{c=1, 2, \cdots, \; C} \; & f_c(\boldsymbol{x}_{adv}) \neq y,  \\
    & \textrm{s.t.} \;  || \boldsymbol{x}_{adv} - \boldsymbol{x} ||_p \leq \epsilon, \; \textrm{and} \; \boldsymbol{x}_{adv} \;, \boldsymbol{x} \; \in \mathcal{D}  \\
    \end{split}
    \label{eq:attack_problem}
\end{equation}
where $\boldsymbol{x}_{adv}$ denotes an adversarial example within the $\epsilon$-ball bounded vicinity of $\boldsymbol{x}$; $||\cdot||_p$ denotes the $\ell_p$ norm where $p$ could be $0,1,2,\infty$. Often, $p$ is selected according to attack algorithms.

\subsection{Adversarial Attack Methods}
Existing adversarial attack methods can be broadly categorized into two types: black-box attacks and white-box attacks \cite{yuan2019adversarial,dong2020benchmarking}. In the black-box attack setting, attackers are assumed to have no knowledge of the victim classifier, but they can be permitted to access it through queries; or attackers may have knowledge of a surrogate classifier. In contrast, in the white-box attack setting, attackers have complete knowledge of the victim classifier (e.g., network architecture, model parameters). Such knowledge is often exploited to craft adversarial examples. In general, white-box attacks succeed more easily than black-box ones, posing significant threats to victim classifiers. 

Different attack methods have been proposed to work effectively in the white-box setting. For example, Goodfellow et al. propose the fast gradient sign method (FGSM) \cite{FGSM}, an efficient single step attack to generate adversarial examples. The projected gradient descent (PGD) attack is an iterative version of FGSM with random starts \cite{madry2018towards}. Carlini and Wagner reformulates Eq. (\ref{eq:attack_problem}) in the Lagrangian form and propose the C\&W attack \cite{carlini2017adversarial}. Momentum FGSM (MIFGSM) introduces the momentum in gradient accumulations at each iteration, which enhances the adversarial transferability \cite{MIFGSM}. 

In the ``arms race'' between attackers and defenders, DCNN classifier owners may employ some defense measure to protect the victim classifier. Correspondingly, attackers can develop defense-aware attacks to evade the victim classifier. In such an attack setting, attackers are assumed to have access to the victim classifier and the defense mechanism. Thus attackers can conduct defense-aware attacks. A common attacking strategy is to fool the victim classifier and its defense jointly in an end-to-end white-box attack manner using gradient-based attacks. Since some defenses may cause the ``obfuscated gradients'' phenomenon (e.g., shattered gradients) \cite{athalye2018obfuscated}, the backward pass differentiable approximation (BPDA) technique is often used to estimate the image gradient in back-propagating gradients \cite{athalye2018obfuscated,carlini2019evaluating}. Moreover, in defense-aware attacks, attackers may try gradient-free black-box attacks to avoid estimating gradients, at the cost of a large number of queries \cite{chen2020hopskipjumpattack}.             

\begin{figure*}[!htbp]
  \centering
  \includegraphics[width=0.75\linewidth]{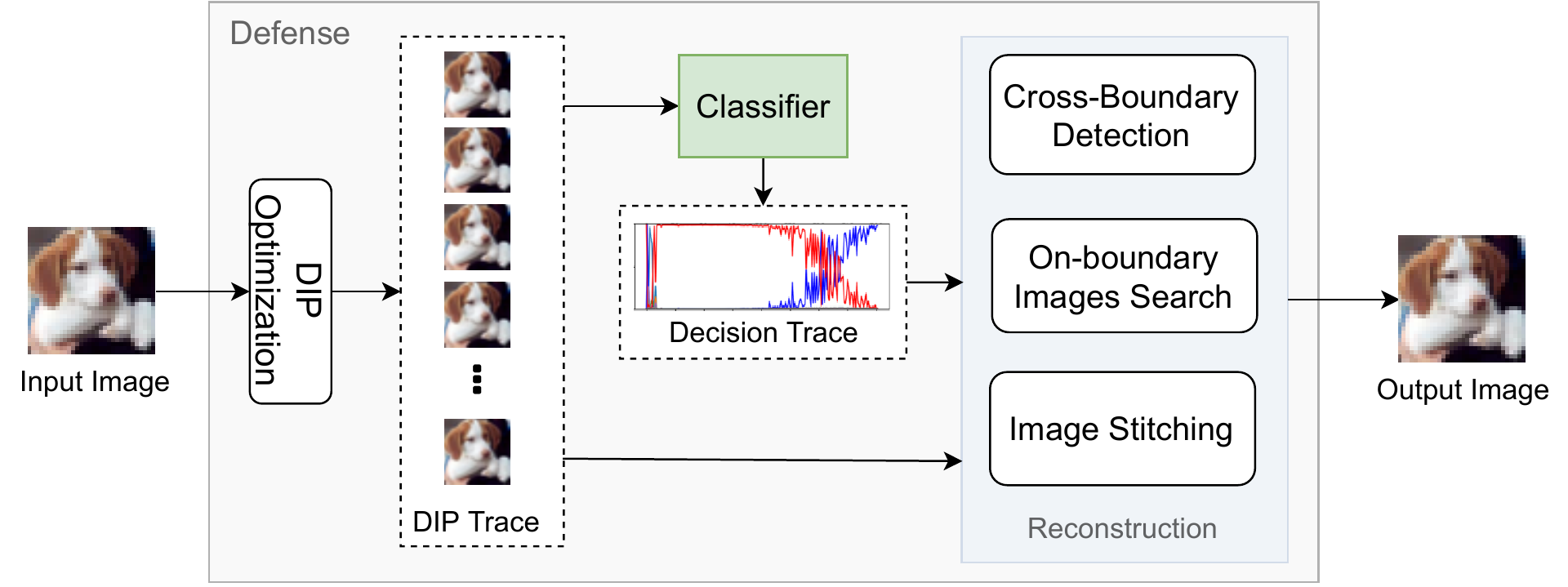}
  \caption{The framework of the proposed reconstruction-based adversarial defense method. Our method firstly maps the DIP trace to the decision space by explicitly incorporating the victim classifier's decisions into defense. A reconstruction module then reliably detects cross-boundary images and search on-boundary images in the decision space. The final reconstructed image can be obtained by stitching on-manifold images.  
  } 
  \label{fig:proposed_framework}
\end{figure*}

\subsection{Adversarial Defense Methods} \label{sec:adv_defenses}
Adversarial defense aims to protect victim classifiers by defending against adversarial attacks. Among existing defense methods, reconstruction-based defenses are gaining increasing popularity. This type of defenses resort to suppressing adversarial noise from adversarial examples. 

Recently, ComDefend \cite{jia2019comdefend}, an end-to-end DCNN compression model, was proposed to remove adversarial noise while simultaneously maintain the structural information of a clean image. Mustafa et al. propose to employ wavelet denoising and a pre-trained super-resolution (SR) network to reconstruct clean images \cite{mustafa2019image}.  Naseer et al. develop a neural representation purifier (NRP) to purify adversarial perturbations in the feature space \cite{naseer2020self}. Despite their effectiveness in defending certain attacks, such defense methods require training on some external datasets (or pretrained models).  

More recently, DIP-based defenses emerge as a type of training-free reconstruction-based defenses \cite{kattamis2019exploring,sutanto2020adversarial,dai2020dipdefend}. By capturing the natural image statistics in a single adversarial image, a DIP network can produce reconstructed images during its optimization. It is likely that some images can be correctly recognized by the victim classifier. Kattamis et al. firstly explore DIP \cite{ulyanov2018deep} as an adversarial defense method. Their preliminary study reveals an interesting phenomenon that the DIP network often overfits to an adversarial image. DIPDefend \cite{dai2020dipdefend} alleviates the overfitting problem utilizing an adaptive stopping criterion, and it achieves the state-of-the-art performance in adversarial defense. Nevertheless,  DIPDefend cannot effectively handle adversarial examples that are near the decision boundary, e.g., adversarial examples generated by the C\&W attack \cite{carlini2017adversarial}.

\section{Methodology}
In this section, we describe a novel and effective reconstruction-based defense against adversarial attacks. The overall framework is illustrated in Fig.~\ref{fig:proposed_framework}. Firstly, we introduce the DIP modeling and analyze the limitation of ``turning point'' selection in the image space used in the state-of-art DIP-based defense \cite{dai2020dipdefend}. We then introduce a novel adversarial defense method that explicitly incorporates the decision process of the victim classifier. As illustrated in Fig.~\ref{fig:proposed_framework}, the proposed method maps each image along the DIP trace to the decision space of the victim classifier. By analyzing the decision process, cross-boundary images can be accurately detected and localized to construct on-boundary images that lie on the decision boundary. Finally, on-boundary images are employed to reconstruct the clean version using the proposed denosing method. 
    
\subsection{DIP Modeling and Analysis}
DIP \cite{ulyanov2018deep} shows that an untrained DCNN can effectively capture the prior of natural images. As a learning-free prior, DIP is often used as an implicit regularizer to solve inverse problems in image reconstruction tasks. Given a single (possibly degenerated) image $\boldsymbol{x_0}$, the reconstruction process seeks a function $g_{\theta}$ which maps a Gaussian noise $\boldsymbol{z}$ to a reconstructed image $\boldsymbol{x}$. The optimization procedure continues until a reconstructed image $\boldsymbol{x}$ is found that satisfies a certain criterion (e.g., high visual quality). Remarkably, even without any training, the DIP framework can achieve comparable image reconstruction performance with data-driven methods that require training on a large dataset.

Suppose that $\boldsymbol{x_0} \in \mathbb{R}^{d}, \; \boldsymbol{z} \in \mathbb{R}^{d'} $, and $g_{\theta}$ is modeled by a DCNN network. Then, the DIP modeling \cite{ulyanov2018deep} is formulated as,
\begin{equation}
    \hat{\theta} = \argmin_{\theta} L( g_{\theta} (\boldsymbol{z}) ; \boldsymbol{x}_0 ), \quad \hat{\boldsymbol{x}} = g_{\hat{\theta}} (\boldsymbol{z})
    \label{eq:dip}
\end{equation}
where $L(\cdot)$ denotes a reconstruction loss function (e.g., the mean-squared-error (MSE)); $\hat{\theta}$ and $\hat{\boldsymbol{x}}$ denote the local optimal parameter and the corresponding reconstructed image, respectively. During DIP optimization, the network parameters $\theta$ are updated at each iteration. Thus a series of reconstructed images can be generated during the optimization process \cite{ulyanov2018deep}.  

Previous works show that reconstructed images from a DIP model (specific to an adversarial example) will pass through the correct decision boundary, and become adversarial again \cite{kattamis2019exploring,sutanto2020adversarial,dai2020dipdefend}. Therefore, it is essential to terminate the DIP reconstruction by correctly localizing the ``turning point'' such that reconstructed images would not become adversarial examples. However, it is challenging to discriminate the location of the decision boundary due to the unavailability of the groundtruth labels. To our best knowledge, DIPDefend \cite{dai2020dipdefend} is the only work that attempts to provide an explicit stopping criterion. DIPDefend assumes that the reconstruction error (between a reconstructed image and an adversarial example) decreases quickly before the ``turning point''. Then, the decreasing tendency becomes slower after this point. DIPDefend \cite{dai2020dipdefend} hypothesizes that the ``turning point'' can be localized by finding the peak of the smoothed peak-to-signal ratio (PSNR) curve, which is computed between an reconstructed image and $\boldsymbol{x}_0$ over iteration number. 

\begin{figure}[!htbp]
  \centering
  \includegraphics[width=0.78\linewidth]{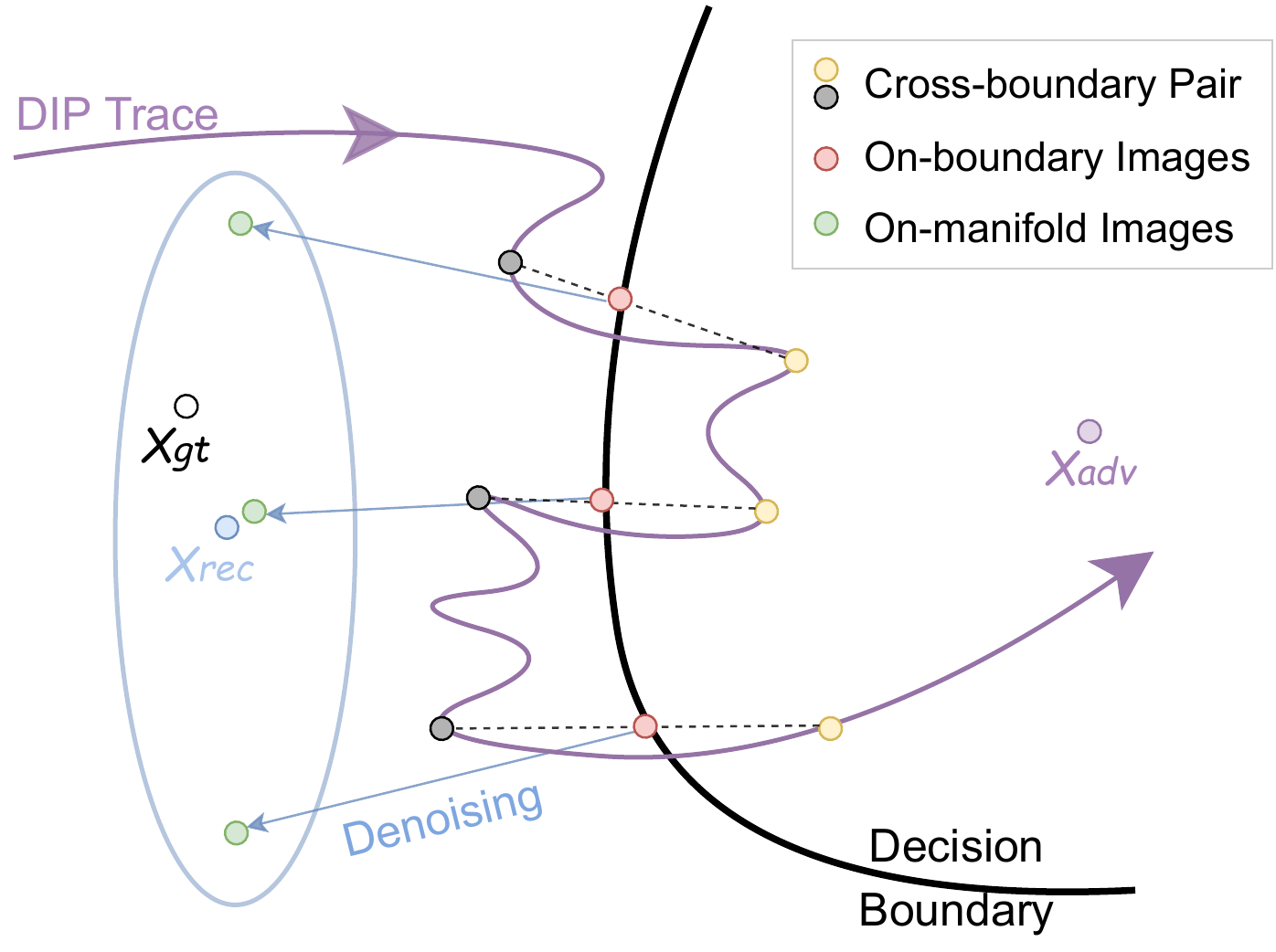}  
  \caption{
  An illustration of the proposed reconstruction module via denoising. Firstly, cross-boundary image pairs are detected and employed to construct on-boundary images. Then, adversarial noise is purified to construct on-manifold images by moving a small step along the reverse direction to the adversarial example $\boldsymbol{x}_{adv}.$ Finally, on-manifold images are stitched to suppress the reconstruction error and obtain a clean image $\boldsymbol{x}_{rec}$.  
  } 
  \label{fig:illustration_cross_boundary}
\end{figure}

Unfortunately, the stopping criterion in \cite{dai2020dipdefend} may capture a wrong ``turning point'' if the adversarial example is near the decision boundary (i.e., only slightly perturbed). This is because the reconstruction process around the decision boundary is smooth which makes it difficult to distinguish whether the reconstructed image contains adversarial noise. Therefore, the ``turning point'' detected via this criterion could be out of the decision boundary which in turn fails to protect the victim classifier.   
Instead of detecting the ``turning point'' by observing the reconstructed image sequence in the image space, we will reconstruct an image by explicitly considering the decision process of the victim classifier. As shown in Fig. \ref{fig:illustration_cross_boundary}, we construct images lying on the decision boundary by detecting cross-boundary images in the decision space. We then purify adversarial noise from the constructed on-boundary images. Finally, we construct the clean version by stitching on-manifold images to remove the reconstruction error in DIP reconstruction.

\subsection{Cross-boundary Images Detection }
This section introduces our method to detect cross-boundary images and further localize on-boundary images. For convenient expression, we denote the DIP reconstructed image sequence as $X=\big \{  \boldsymbol{x}^{(t)}  \big \}_{t=0}^{T-1}$. The victim classifier $f$ predicts labels of $X$ as $Y=\big \{\argmax_{c=1,2,\cdots,\;C} f_c ( \boldsymbol{x}^{(t)} ) \big \}_{t=0}^{T-1}$. In cross-boundary images detection, the objective is to find reconstructed images that satisfy two constraints: C1) the reconstructed image should be perceptually recognizable so that it could be correctly predicted by the victim classifier; and C2) adjacent reconstructed images should cross the decision boundary so that we could possibly reconstruct images on the decision boundary. Reconstructed images that satisfy both C1) and C2) are termed as cross-boundary images. Formally, these images can be detected using the formula,
\begin{equation}
    I = \big \{ t \; | \; y^{(t)} \neq y^{(t+1)}  \land v^{(t)} > \tau \land t > t_0   \big \}
    \label{eq:cross_bd_detection}
\end{equation}
where $v^{(t)}$ denotes the perceptual quality of the reconstructed image at iteration $t$, $\tau$ denotes the perceptual quality threshold, and $t_0$ denotes a warmup iteration number in DIP optimization. Due to the unavailability of the groundtruth clean image, $v^{(t)}$ can be approximately computed by comparing $\boldsymbol{x}^{(t)}$ and the given image $\boldsymbol{x}_0$. Considering the perceptual alignment with the human visual system, we recommend using the structural similarity (SSIM) \cite{wang2004image} metric, then $v^{(t)} \approx \textrm{SSIM} \big (\boldsymbol{x}^{(t)}, \boldsymbol{x}_0 \big)$.

Next, we construct images that lie on the decision boundary with our detected cross-boundary images. For $\forall t\in I$, we assume the existence of $\boldsymbol{x}_{db}^{(t)}$ that can be linearly interpolated based on cross-boundary images $\boldsymbol{x}^{(t)}$ and $\boldsymbol{x}^{(t+1)}$, 
\begin{equation}
    \boldsymbol{x}_{db}^{(t)} = \alpha \boldsymbol{x}^{(t)} + (1-\alpha) \boldsymbol{x}^{(t+1)}, \; t \; \in I 
    \label{eq:boundary_images}
\end{equation}
where $\alpha \in [0,1]$. An optimal parameter $\alpha^{opt}$ can be obtained that satisfies,
\begin{equation}
    \alpha^{opt} = \argmin_{\alpha} \big | f_{y^{(t)}} ( \boldsymbol{x}_{db}^{(t)}  )  - f_{y^{(t+1)}} ( \boldsymbol{x}_{db}^{(t)}  )   \big | , \; t \; \in I
\end{equation}
Here we use the grid search to find a suitable $\alpha$. Suppose that we equally separate the grid into $N$ sections, the error between $\boldsymbol{x}_{db}^{(t)}$ and its true value is bounded by $\frac{1}{2N} || \boldsymbol{x}^{(t)} - \boldsymbol{x}^{(t+1)} || $. Clearly, we can always reduce the approximation error by setting a larger $N$.  

It is worth noting that cross-boundary images may densely locate around the decision boundary. In such a circumstance, for computationally efficiency, we only need to use a subset of cross-boundary images, e.g., $I_K \subseteq I$, where $I_K$ denotes the last $K$ cross-boundary images that are detected along the DIP trace. Conversely, if no cross-boundary images are detected, it indicates that the DIP trace has not passed through the decision boundary. In this case, we simply keep the last $K$ reconstructed images $\big \{\boldsymbol{x}^{(t)} \big \}_{t=T-K}^{T-1}$ as our detected images.          
\begin{algorithm}[tb]
	\footnotesize
	\SetAlgoLined
	\KwData{An input image $\boldsymbol{x}\in\mathbb{R}^d$, a Gaussian noise $\boldsymbol{z}\in\mathbb{R}^{d'}$, pre-trained victim classifier $f$, maximum iteration number $T$, warmup iteration $t_0$, SSIM threshold $\tau$, maximum number of cross-boundary images $K$, stepsize $\beta$.} 
 	\KwResult{A reconstructed image $\hat{\boldsymbol{x}}_{rec}$.}   
	Initialize a DIP network $g_{\theta^{(0)}}$ using xavier initialization, initialize DIP image set $X=\phi$, initialize classifier decision set $Y=\phi$, initialize cross-boundary image index set $I=\phi$, initialize on-boundary image set $X_{db}$\;
 	\tcp{\textbf{Generate and map DIP image sequence to decision space}}  
	\For{$t=0$ \KwTo $T-1$}{   
	Input $\boldsymbol{z}$ to DIP network $g_{\theta}^{(t)}$, and output $\boldsymbol{x}^{(t)} \leftarrow g_{\theta}^{(t)}(\boldsymbol{z})$\; 
	Input $\boldsymbol{x}^{(t)}$ to $f$, and output label $y^{(t)} \leftarrow \argmax_{c=1,2,\cdots,\; C} f_c ( \boldsymbol{x}^{(t)} )$\;
	Update DIP image sequence: $X\leftarrow X \cup \boldsymbol{x}^{(t)}$\;
	Update classifier decision set: $Y\leftarrow Y \cup y^{(t)}$\;
	Compute $\textrm{MSE}(\boldsymbol{x}^{(t)}, \boldsymbol{x})$, and compute $g_{\theta^{(t+1)}}$ from $g_{\theta^{(t+1)}}$ with gradient descent optimization\;
	}
	\tcp{\textbf{Cross-boundary images detection}}
	\For{$t=t_0$ \KwTo $T-1$}{
	Compute $v^{(t)}: v^{(t)} \leftarrow \textrm{SSIM}(\boldsymbol{x}^{(t)}, \boldsymbol{x})$\;
	\If{$y^{(t)} \neq y^{(t+1)}$  \&  $v^{(t)} \geq \tau$ }{
	Update cross-boundary image index set: $I \leftarrow I \cup t $\; 
	Localize on-boundary image $\boldsymbol{x}_{db}^{(t)}$ from Eq. (\ref{eq:boundary_images})\;
	Update on-boundary image set: $X_{db} \leftarrow X_{db} \cup \boldsymbol{x}_{db}^{(t)}$\;
	}
	}
	\tcp{\textbf{On-manifold images stitching}}
	Construct on-manifold images $\boldsymbol{x}_{manifold}^{(t)}$ for $\in I_K$ from Eq. (\ref{eq:manifold_perturb})\;
	Obtain a clean image $\hat{\boldsymbol{x}}_{rec}$ from Eq. (\ref{eq:final_rec_img})\;
	\textbf{Return} $\hat{\boldsymbol{x}}_{rec}$\;
	\caption{The proposed reconstruction-based algorithm for adversarial defense.}
	\label{alg:CSM}
\end{algorithm}

\subsection{On-manifold Images Stitching}
After constructing on-boundary images, our next objective is to reconstruct a clean image that can be correctly predicted by the victim classifier. The reconstruction goal is realized through the proposed two-step denoising method: 1) Purifying adversarial noise by pushing on-boundary images back to the manifold; and 2) Suppressing reconstruction errors by stitching on-manifold images to construct a clean image.  

The first step is to slightly perturb on-boundary images such that they are pushed back to the correct manifold. Intuitively, the perturbation direction should be along with the gradient of the on-boundary image. However, the gradient may also direct to the adversarial image manifold. Alternatively, we move a small step along the reverse direction of the adversarial noise, i.e., $\boldsymbol{d^{(t)}}=\boldsymbol{x}_{bd}^{(t)} - \boldsymbol{x}_0, \; t\in I_K$. The on-manifold images can be obtained by,
\begin{equation}
    \boldsymbol{x}_{manifold}^{(t)} = \boldsymbol{x}_{bd}^{(t)} + \beta \boldsymbol{d^{(t)}}, \; t\in I_K 
    \label{eq:manifold_perturb}
\end{equation}
where $\boldsymbol{x}_{manifold}^{(t)}$ denotes a on-manifold image, and $\beta \geq 0$ denotes the perturbation stepsize of the on-boundary image $\boldsymbol{x}_{bd}^{(t)}$. Specially, we set $\beta=0$ if $I_K=\phi$, since these detected images have already been on the correct manifold.     

Finally, we craft the reconstructed image $\hat{\boldsymbol{x}}_{rec}$ as,
\begin{equation}
    \hat{\boldsymbol{x}}_{rec} = \frac{1}{K} \sum_{k=1}^K \boldsymbol{x}_{manifold}^{(t)} \;, \; t\in I_K 
    \label{eq:final_rec_img}
\end{equation}

In Algorithm 1, we show the details of the proposed method.

\section{Experiments}
In this section, we will demonstrate the superiority of the proposed reconstruction-based scheme in defending against adversarial attacks. We first introduce our experimental setup, and then conduct comparisons with state-of-the-art defense methods. Experimental results show that the proposed method consistently outperforms baseline defense methods with a large margin, for both white-box attacks and defense-aware attacks.

\begin{table*}[!htbp]
\caption{Performance comparisons with five adversarial defense methods on the CIFAR-10 and ImageNet datasets under white-box non-targeted attacks. Each entry shows the model accuracy (\%) under an attack/defense combination. FGSM \cite{FGSM}, PGD \cite{madry2018towards}, BIM \cite{Iteradv17} and MIFGSM \cite{MIFGSM} attacks adopt the $\ell_{\infty}$ norm with $\epsilon=2/4/8$. C\&W \cite{carlini2017adversarial} and DDN \cite{rony2019decoupling} attacks adopt the $\ell_{2}$ norm with default parameters. The best performances have been marked in bold.}
\label{tab: quant}
\centering
\begin{adjustbox}{width=0.999\textwidth}
\begin{tabular}{lcccccccc|c} 
\toprule
Dataset &Defense Method &Clean &FGSM \cite{FGSM} &PGD \cite{madry2018towards} &BIM \cite{Iteradv17} &MIFGSM \cite{MIFGSM} &C\&W \cite{carlini2017adversarial} &DDN \cite{rony2019decoupling} & $\overline{\textrm{ACC}}$ \\\midrule
&No defense &92.9 &57.5/49.2/44.4 &24.6/1.9/0 &49.8/19.2/2.4 &46.7/29.7/11.5 &0.9 &0.4 &28.7 \\
&ComDefend \cite{jia2019comdefend} &89.5 &72.6/60.5/51.0	&75.7/62.0/50.1	&70.5/57.0/36.7	&69.3/51.1/30.4	&\textbf{86.4}	&57.4	&61.3\\
CIFAR-10&SR \cite{mustafa2019image} &45.2 &44.2/42.1/37.7 &41.4/36.8/30.3 &43.4/40.0/33.0 &43.8/40.5/34.4 &35.3 &33.5 &38.8 \\
&NRP \cite{naseer2020self} &\textbf{90.9} &65.1/56.8/57.7 &58.4/35.1/35.9 &61.5/31.2/13.0 &59.4/39.4/23.1 &30.5 &15.5 &44.9 \\
&DIPDefend \cite{dai2020dipdefend} &83.2 &78.6/72.6/58.4 &79.0/77.6/73.6 &79.2/72.0/67.6 &76.2/69.4/51.6 &71.2 &76.6 &71.5 \\
&\textbf{Proposed} &85.6 &\textbf{80.4}/\textbf{79.8}/\textbf{60.1} &\textbf{87.5}/\textbf{86.9}/\textbf{83.1} &\textbf{84.6}/\textbf{86.7}/\textbf{76.1} &\textbf{83.4}/\textbf{79.9}/\textbf{60.4} &80.8  &\textbf{83.4} &\textbf{79.9} \\
\midrule
&No defense &90.3 &6.7/5.2/5.8 &0/0/0 &0.8/0.1/0 &0.2/0.1/0.1 &0 &1 &7.35 \\
&ComDefend \cite{jia2019comdefend} &81.1	&34.2/19.7/12.7	&42.2/18.4/2.8	&30.2/11.7/3.8	&23.9/4.8/0.6	&77.8	&63.1	&28.4 \\
ImageNet&SR \cite{mustafa2019image} &73.3 &64.0/56.6/\textbf{43.7} &68.1/63.4/49.3 &65.6/59.1/53.3 &63.7/53.5/\textbf{38.2} &70.7 &70.5 &59.5 \\
&NRP \cite{naseer2020self} &\textbf{86.7} &32.4/21.4/18.5 &38.4/16.9/13.3 &26.5/7.1/2.2 &18.4/2.8/3.3 &82.1 &59.5 &28.6 \\
&DIPDefend \cite{dai2020dipdefend} &73.0 &60.8/47.2/32.7 &65.5/61.3/49.1 &58.3/54.2/48.7 &56.8/45.5/32.9 &70.5 &68.9 &55.0 \\
&\textbf{Proposed} &84.5 &\textbf{74.2}/\textbf{58.7}/39.4 &\textbf{80.6}/\textbf{80.1}/\textbf{67.8} &\textbf{79.8}/\textbf{76.6}/\textbf{61.2} &\textbf{75.4}/\textbf{54.1}/33.7 &\textbf{82.7} &\textbf{82.8} &\textbf{68.7} \\

\bottomrule
\end{tabular}
\end{adjustbox}
\end{table*}

\subsection{Experimental Setup} \label{sec:exp_setup}
\textbf{Datasets:} We conduct experiments on CIFAR-10 \cite{krizhevsky2009learning} and ImageNet \cite{deng2009imagenet} subsets. In CIFAR-10, we use a pre-trained ResNet-50 \cite{he2016deep} as the victim classifier, whereas a pre-trained ResNet-101 is employed as the victim classifier on ImageNet. For evaluation, on CIFAR-10, we randomly sample 1000 images from the test set and the image size is $32\times32$. On ImageNet, we use the 1000 image samples officially provided by the NeurIPS 2017 competition track on non-targeted adversarial attacks \cite{kurakin2018adversarial}. The ImageNet samples that we evaluate are resized to be $224\times224$.    
 
{\setlength{\parindent}{0cm}\textbf{Attack and Defense Methods:}} For attacks, an attacker first performs \textit{white-box attacks}. In this setting, assume that they can have full access to victim classifiers to conduct white-box attacks; however, classifier owners may additionally perform defenses to counter such attacks. These defenses are assumed unknown to the attacker. Specifically, we assume attacker adopts five dominant gradient-based attacks: FGSM \cite{FGSM}, PGD \cite{madry2018towards}, BIM \cite{Iteradv17}, MIFGSM \cite{MIFGSM}, C\&W \cite{carlini2017adversarial} and DDN \cite{rony2019decoupling}. In this attack setting, classifier owners will employ and evaluate the effectiveness of defense methods to protect victim classifiers. The defense methods include three recently proposed training-based defense methods: ComDefend \cite{jia2019comdefend}, SR \cite{mustafa2019image}, NRP \cite{naseer2020self}. Also, the DIPDefend \cite{dai2020dipdefend}, a training-free defense yet achieves the state-of-the-art defense performance, is evaluated and compared with the proposed method. Please refer to Section~\ref{sec:adv_defenses} for brief introduction of attack/defense methods to be evaluated.

In addition, we evaluate the defense methods under \textit{defense-aware attacks} \cite{carlini2019evaluating}. In this setting, adversaries are assumed to be aware of and have full knowledge to both victim classifiers and defense mechanisms adopted by classifier owners. Defense-aware attackers may re-design attacks to fool victim classifiers that are protected by defense methods. As suggested in \cite{carlini2019evaluating}, attackers generally adopt BPDA-based white-box attacks \cite{athalye2018obfuscated} and gradient-free black-box attacks through queries, e.g. \cite{chen2020hopskipjumpattack}.

{\setlength{\parindent}{0cm}\textbf{Parameters and Metrics:}} The parameters of attack methods are chosen to significantly degrade the performance of victim classifiers. Specifically, for FGSM \cite{FGSM}, PGD \cite{madry2018towards}, BIM \cite{Iteradv17}, MIFGSM \cite{MIFGSM} attack methods, we adopt the $\ell_{\infty}$ norm with the perturbation bound $\epsilon$ as 2, 4 and 8, respectively. C\&W \cite{carlini2017adversarial} and DDN \cite{rony2019decoupling} attacks adopt the $\ell_{2}$ norm with default parameters as in official implementations. Similarly, we adopt the officially implemented defense methods or pre-trained models (e.g., ComDefend \cite{jia2019comdefend}, NRP \cite{naseer2020self}) with their default parameters. For the proposed method, we use the U-Net network architecture with skip connection \cite{ronneberger2015u} as the DIP model following the official DIP implementation in \cite{ulyanov2018deep}. The DIP model is initialized with random noise, and it applies MSE as the loss function.
The iteration number $T$ are set to $1000$ for experiments on CIFAR-10 and $3000$ on ImageNet. For calculation efficiency, we map 200 images at equal intervals along DIP trace to their decision space.
In the detection module, we set $\tau=0.7, t_0=50$ and $K=15$. In the reconstruction module, we set $\beta=0.5$. In evaluation, we adopt classification accuracy as the performance metric. Ideally, an effective defense method can maintain high classification accuracy for both clean images and adversarial examples. Please refer to Appendix A for details of our experimental setup.

\subsection{Evaluation Results on White-box Attacks} \label{sec:eval_white_box}  
In Section~\ref{sec:quant_white_box}, we first quantitatively evaluate the defense effectiveness of the proposed method against white-box attacks on CIFAR-10 and ImageNet. Then, in Section~\ref{sec:visual_white_box}, we compare the image quality of reconstructed images from five different defense methods.

\subsubsection{Defense Results Comparison} \label{sec:quant_white_box}
Table~\ref{tab: quant} reports the defense result comparisons under white-box attacks on CIFAR-10 and ImageNet, respectively. For fair comparison, we also report accuracy comparison on clean images which are not attacked. Also, for reference, we show in the ``No defense'' row the accuracy of victim classifiers without adopting a defense method.

\begin{table*}[!htbp]
\caption{Image quality comparisons of reconstructed images from different adversarial defense methods. Here we adopt the SSIM \cite{wang2004image} metric. The best performances have been marked in bold.}
\label{tab: iqa}
\centering
\begin{adjustbox}{width=0.999\textwidth}
\begin{tabular}{lcccccccc|c} 
\toprule
Dataset &Defense Method &Clean &FGSM \cite{FGSM} &PGD \cite{madry2018towards} &BIM \cite{Iteradv17} &MIFGSM \cite{MIFGSM} &C\&W \cite{carlini2017adversarial} &DDN \cite{rony2019decoupling} &$\overline{\textrm{SSIM}}$ \\\midrule
&ComDefend \cite{jia2019comdefend} &0.925	&0.922/0.912/0.877	&0.922/0.916/0.890	&0.923/0.919/\textbf{0.907}	&0.922/0.914/\textbf{0.891}	&0.924	&0.915	&0.912\\
&SR \cite{mustafa2019image} &0.901 &0.898/0.887/0.830 &0.898/0.891/0.860 &0.899/0.894/0.879 &0.898/0.891/0.856 &0.900 &0.889 &0.884 \\
CIFAR-10 &NRP \cite{naseer2020self} &0.884 &0.881/0.879/0.888 &0.882/0.879/0.880 &0.882/0.879/0.873 &0.881/0.877/0.872 &0.883 &0.872 &0.879 \\
&DIPDefend \cite{dai2020dipdefend} &0.919 &0.909/0.884/0.811 &0.912/0.895/0.849 &0.911/0.899/0.874 &0.912/0.895/0.851 &0.907 &0.897 &0.888 \\
&\textbf{Proposed} &\textbf{0.943} &\textbf{0.941/0.934/0.913} &\textbf{0.930/0.918/0.891} &\textbf{0.935/0.922/0.907} &\textbf{0.940/0.915}/0.868 &\textbf{0.941} &\textbf{0.951} &\textbf{0.923} \\
\midrule
&ComDefend \cite{jia2019comdefend} &0.862	&0.855/0.832/0.764	&0.857/0.847/0.815	&0.858/0.853/0.843	&0.857/0.843/0.793	&0.861	&0.861	&0.840 \\
&SR \cite{mustafa2019image} &0.885 &0.880/0.861/0.762 &0.882/0.874/0.842 &0.882/0.879/0.869 &0.881/0.871/0.813 &0.884 &0.884 &0.863 \\
ImageNet &NRP \cite{mustafa2019image} &0.823 &0.825/0.830/\textbf{0.869} &0.823/0.823/0.844 &0.826/0.825/0.825 &0.826/0.821/0.835 &0.827 &0.826 &0.830 \\
&DIPDefend \cite{dai2020dipdefend} &0.802 &0.789/0.746/0.640 &0.795/0.771/0.709 &0.794/0.782/0.763 &0.795/0.771/0.696 &0.807 &0.803 &0.764 \\
&\textbf{Proposed} &\textbf{0.926} &\textbf{0.898/0.867/}0.839 &\textbf{0.925/0.891/0.845} &\textbf{0.905/0.891/0.889} &\textbf{0.901/0.892/0.851} &\textbf{0.923} &\textbf{0.940} &\textbf{0.892} \\
\bottomrule
\end{tabular}
\end{adjustbox}
\end{table*}

In Table~\ref{tab: quant}, we observe that ResNet-50 achieves a high classification accuracy (i.e., 92.9\%) on CIFAR-10 without adversarial attacks. However, its performance drops significantly in the presence of adversarial attacks. Particularly for PGD ($\epsilon=8$) \cite{madry2018towards}, C\&W \cite{carlini2017adversarial}, and DDN \cite{rony2019decoupling}, the classification accuracy of ResNet-50 drops close to 0, indicating that such three attacks poses highest threats to the victim classifier in this setting. Next, we find that all evaluated defenses can mitigate the effects of adversarial examples. For example, DIPDefend \cite{dai2020dipdefend}, the state-of-the-art method, greatly enhances the classification accuracy of adversarial examples. Nevertheless, the proposed method further significantly improves over all existing baseline methods except C\&W. For example, our method outperforms the state-of-the-art method (i.e., DIPDefend \cite{dai2020dipdefend}) by \textbf{9.5}\%/\textbf{9.6}\%/\textbf{6.8}\% on PGD ($\epsilon=8$) \cite{madry2018towards}/C\&W \cite{carlini2017adversarial}/DDN \cite{rony2019decoupling} attacks, respectively. Moreover, we observe that the proposed method also maintains a high classification accuracy (i.e., 85.6\%) on clean images. This observation indicates that the our method can effectively recover the essential information of clean images for correct classification. It is worthy mentioning that, although NRP \cite{naseer2020self} and ComDefend \cite{jia2019comdefend} achieve a higher accuracy than our method on clean images, our method significantly outperforms them in terms of the averaged accuracy denoted by $\overline{\textrm{ACC}}$. Specifically., NRP \cite{naseer2020self} obtains an $\overline{\textrm{ACC}}$ as 44.9\% where we achieve 79.9\%, which is 35\% higher than the NRP method.  

On ImageNet, this experiment adopts ResNet-101 as the victim classifier which achieves a 90.3\% accuracy without defense on clean image samples. Compared with results on CIFAR-10 in the ``No defense'' situation, we find that adversarial examples fool the victim classifier more easily on ImageNet in white-box attacks. For example, even with $\epsilon=2$, the PGD \cite{madry2018towards} attack can completely fool the victim classifier. This  observation in turn suggests it is even more challenging to defend adversarial examples on the ImageNet dataset. Nevertheless, as a training-free method, the proposed method still achieves an $\overline{\textrm{ACC}}$ as 68.7\%, which is \textbf{9.2}\% and \textbf{13.7}\% higher than the second (i.e., SR \cite{mustafa2019image} which requires training) and third (i.e., DIPDefend \cite{dai2020dipdefend}, which is training-free) most effective defense methods. Moreover, the proposed method also maintains a high accuracy (i.e., 84.5\%) for clean samples, which is comparable to the best performance (i.e., 86.7\%) achieved by NRP \cite{naseer2020self}.

\vspace{-2mm}
\subsubsection{Image Quality Assessment} \label{sec:visual_white_box}
In addition to defense capability comparison, we compare the image quality of reconstructed images from different reconstruction-based defense methods. Table~\ref{tab: iqa} shows the quantitative comparison results on CIFAR-10 and ImageNet using the SSIM metric \cite{wang2004image}. For completeness, we compute and report the visual quality comparison among five evaluated defenses under six types of adversarial attacks. We observe that the proposed method generally outperforms baseline methods with a large margin in terms of image visual quality. Particularly for the defense setting under DDN attacks \cite{rony2019decoupling}, the proposed method outperforms the second best method by \textbf{3.6}\% on CIFAR-10 (achieved by ComDefend \cite{jia2019comdefend})  and \textbf{5.6}\% (achieved by SR \cite{naseer2020self}) on ImageNet, respectively. The superiority of reconstructed images also confirms the effectiveness of the proposed two-stage denoising strategy in improving the image quality. We also show some visual examples in Appendix C. 

\subsection{Evaluation Results on Defense-Aware Attacks} \label{sec:eval_secondary}
This section evaluates and compares the performance of defense methods against defense-aware attacks where attackers are aware of the defenses. Following \cite{athalye2018obfuscated,carlini2019evaluating}, we consider the BPDA-based white-box attack \cite{athalye2018obfuscated} and query-based black-box attack respectively as defense-aware attacks.  

\subsubsection{BPDA-based Attacks}  BPDA-based attack is a special type of white-box attack where attackers have full knowledge of both the victim classifier and its defense mechanism. Let $F(\boldsymbol{x}) \triangleq g \circ f (\boldsymbol{x})$, where $F$ denotes the end-to-end composite function of the defense $g$ and victim classifier $f$. The BPDA technique \cite{athalye2018obfuscated} is adopted when the defense $g$ includes one or more non-differentiable components. In BPDA-based attack, an attacker can approximately estimate the image gradient $\nabla_{\boldsymbol{x}} F(\boldsymbol{x})$ with a standard forward pass through $g \circ f (\boldsymbol{x})$, but replacing $g$ with an approximate but differentiable function $\tilde{g}$ in the backward pass. Specially, attackers can simply set $\tilde{g}$ as an identity mapping function in reconstruction-based defenses. After obtaining the gradient estimate, attackers can craft adversarial examples using the PGD attack \cite{madry2018towards}. 

\begin{figure}[!htbp]
  \centering
  \includegraphics[width=0.98\linewidth]{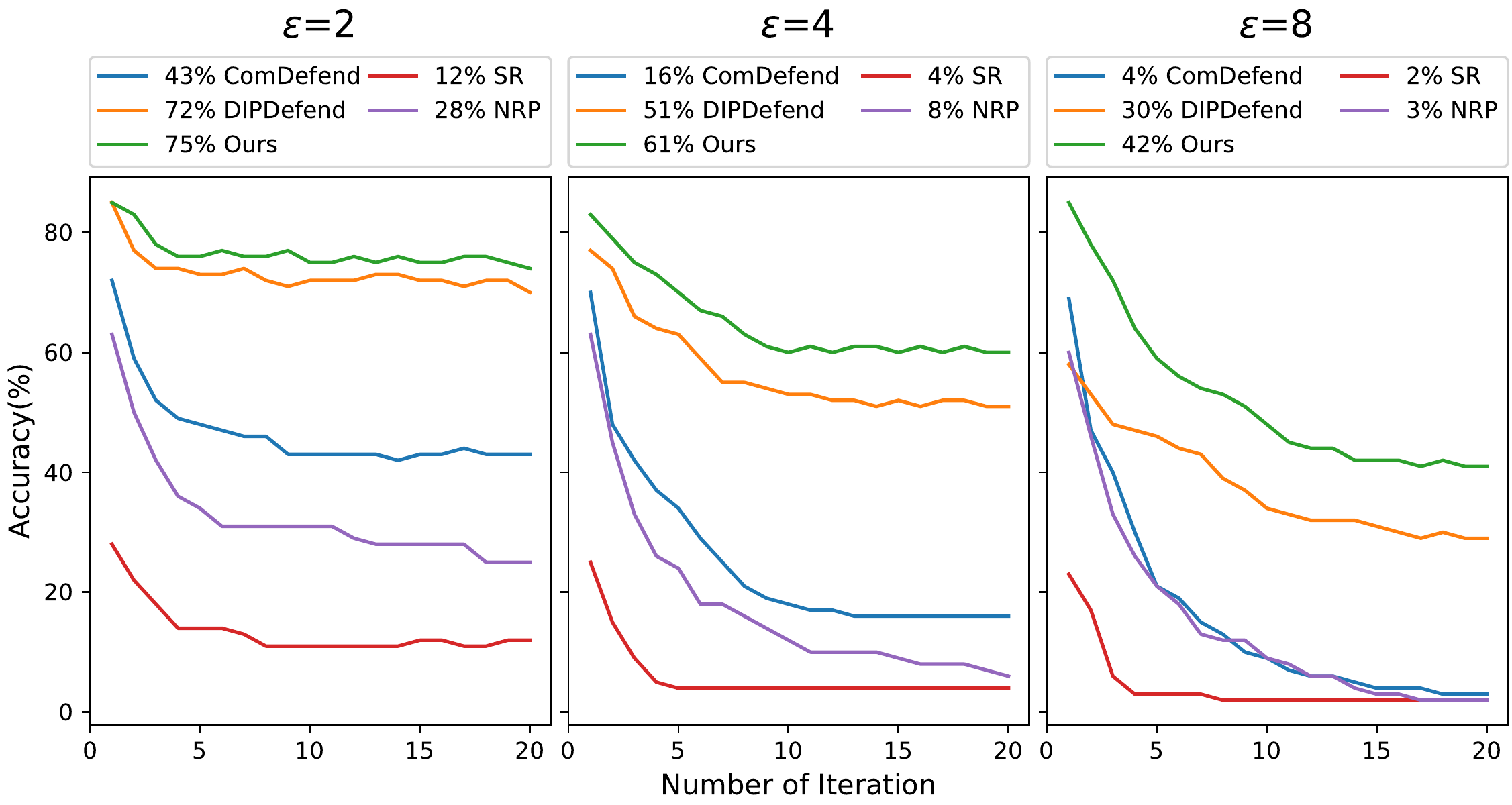}
  \caption{Robustness curves under BPDA-based attacks on CIFAR-10. Three sub-figures show the accuracy comparisons of five defense methods under BPDA-based attacks with $\ell_{\infty}$ as 2 (left), 4 (middle) and 8 (right), respectively. The reported accuracy (in boxes) sets iteration number as 15.
  } 
  \label{fig:bpda_curve}
\end{figure}

In Fig. ~\ref{fig:bpda_curve}, we show the comparison results of five defense methods under BPDA-based attacks with varying perturbation budgets, i.e., $\ell_{\infty}=2/4/8$, respectively. We observe that the proposed method displays robustness to BPDA attack when the perturbation budget is $\ell_{\infty}=2$. By contrast, the accuracy of all baseline defenses, except the DIPDefend \cite{dai2020dipdefend}, degrade significantly. With the number of iteration increases, the proposed method also outperforms DIPDefend \cite{dai2020dipdefend} by about 3\%. When the perturbation budget increases (e.g., $\ell_{\infty}=4/8$), the proposed method achieves substantially higher accuracy than all baseline methods. For example, our method outperforms the baseline defenses by about \textbf{10}\% (i.e., DIPDefend \cite{dai2020dipdefend}) to \textbf{50}\% (i.e., SR \cite{mustafa2019image}). This is because baseline defenses (i.e., SR \cite{mustafa2019image}, ComDefend \cite{jia2019comdefend}, NRP \cite{naseer2020self}) are not particularly designed for BPDA-based attacks.

\subsubsection{Query-based Attacks}
\begin{figure}[!htbp]
  \centering
  \includegraphics[width=0.98\linewidth]{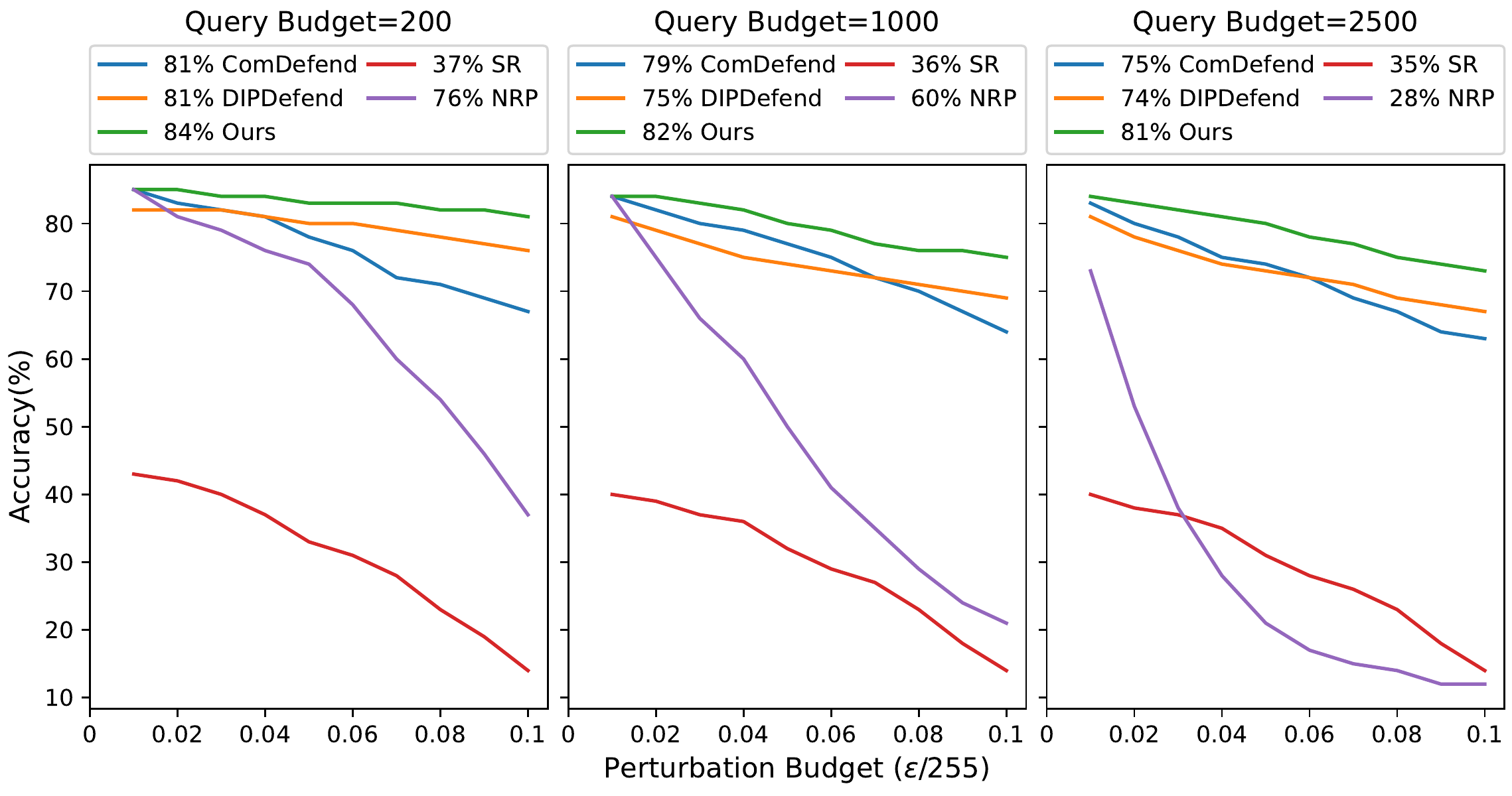}
  \caption{Robustness curves under HSJ attacks \cite{chen2020hopskipjumpattack} on CIFAR-10. Three subfigures show the accuracy comparison of defense methods under HSJ attacks with
  query budget as 200 (left), 1000 (middle) and 2500 (right), respectively. 
  The reported accuracy (in boxes) uses $\epsilon=8$. 
  } 
  \label{fig:hsj_curve}
\end{figure}

We also evaluate our defense against black-box attacks through queries which are agnostic to defense methods.  In experiments, we adopt the HopSkipJumpAttack (HSJ) \cite{chen2020hopskipjumpattack} attack, a state-of-the-art decision-based black-box attack. Fig. ~\ref{fig:hsj_curve} shows the comparison results of five defenses under different query budgets (i.e., 200, 1000, 2500) and perturbation budgets (i.e., $\epsilon \in [0, 25]$) on CIFAR-10. A larger query number (or higher perturbation budget) gives a stronger attack. In general, we observe our method displays robustness against HSJ attacks, particularly for an acceptable query budget (e.g., within 1000). Even if the victim classifier allows a large number queries as 2500, the proposed method still consistently outperforms baseline methods. The vulnerability of some defenses (e.g., SR \cite{naseer2020self}) against HSJ attack may be because they are not designed for such attacks.

\subsection{Evaluations on Parameter Sensitivity}
\label{sec:param_sen}
This section studies the parameter sensitivity of our method, which involves two groups of parameters. The first group includes the perceptual quality threshold $\tau$, the warmup iteration number $t_0$, the number of on-boundary images $K$ and stepsize $\beta$ in the reconstruction module; while the second group contains the iteration number $T$ and network depth during DIP optimization. We study and report the effects of these parameters on CIFAR-10.    

\begin{figure}[!htbp]
  \centering
  \includegraphics[width=0.6\linewidth]{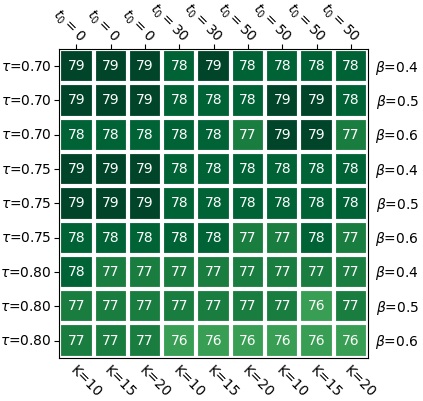} 
  \caption{Average performance comparison of our method under different attacks using combinations of parameters in the first group ($\epsilon=8$). 
  } 
  \label{fig:paraSen1}
\end{figure}

\begin{figure}[!htbp]
  \centering
  \includegraphics[width=0.98\linewidth]{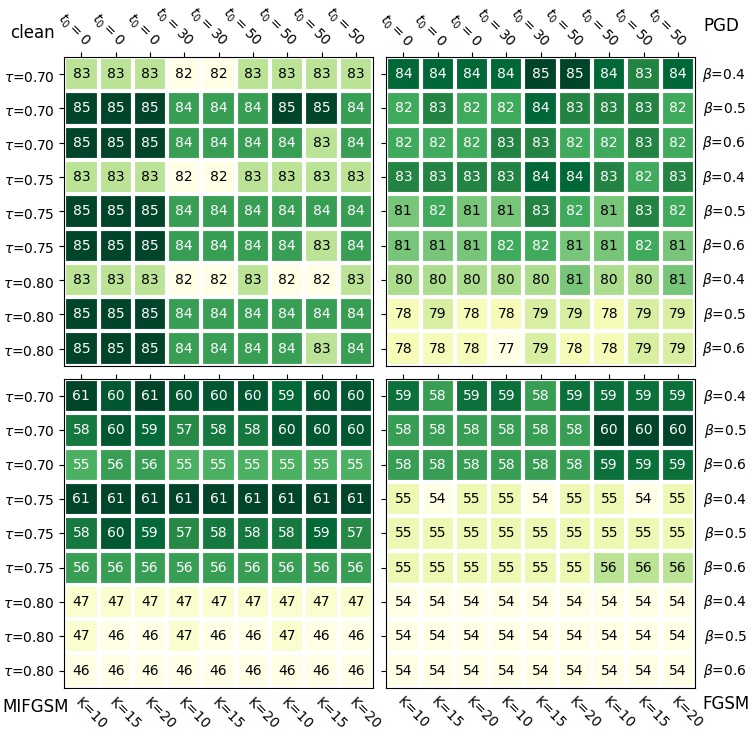}
  \caption{Performance comparison of our method under four different attacks using combinations of parameters in the first group ($\epsilon=8$).} 
  \label{fig:paraSen2}
\end{figure}

For the first group of parameters, Fig.~\ref{fig:paraSen1} reports the average accuracy (i.e., $\overline{\textrm{ACC}}$) of the proposed method under different attacks. Despite with different combinations of parameter, our method maintains a stable defense performance, i.e., $\overline{\textrm{ACC}}$ ranges from 76\% to 79\%. Fig.~\ref{fig:paraSen2} reports detailed performances of different parameter combinations on reconstructing clean/attacked images, where we have a similar observation. These results imply that our method is generally not sensitive to parameter selection.           

Table \ref{tab: paraSen2} reports the average performance comparison of various parameter combination pairs from the second group. 
We compare three networks, i.e., small/medium/large-sized networks (see Appendix B for architecture details). We find that our method is generally robust to different network architectures and iteration number $T$, in particular, a medium-sized network with $T$ equals 1000 (or 750) yields the best defense performance.        

\begin{table}[!htp]
\caption{Performance comparison of our method with varying iteration number $T$ and small/medium/large network architectures. The attack method is PGD \cite{madry2018towards} ($\epsilon=8$). }
\label{tab: paraSen2}
\begin{adjustbox}{width=0.8\linewidth}
\begin{tabular}{lccccc}\toprule
Network &$T$=500 &$T$=750 &$T$=1000 &$T$=1250 &$T$=1500 \\\midrule
small  &76 &74 &74 &72 &69 \\
medium  &75 &79 &79 &75 &74 \\
large   &71 &73 &71 &69 &67 \\
\bottomrule
\end{tabular}
\end{adjustbox}
\end{table}

\section{Conclusion}
In this work, we present a novel training-free and attack-agnostic reconstruction-based defense framework by delving into deep image prior (DIP). Our method efficiently resolves the overfitting problem in DIP optimization. Specifically, instead of analyzing the DIP trace empirically in image space, this work explicitly incorporates the victim classifier into defense and analyzes the DIP trace in decision space. In the decision space, we can reliably detect cross-boundary images and construct on-boundary images. We then develop a novel denoising method where adversarial noise can be efficiently purified from on-boundary images, and reconstruction error can be further suppressed. Extensive experiments on public image datasets show the superiority of the proposed method in defending white-box attacks and defense-aware attacks.

 

\section{Acknowledgments}
We acknowledge financial support from the National Natural Science Foundation of China (NSFC) under Grant No. 61936011, and the Natural Sciences and Engineering Research Council of Canada  (NSERC).


\vspace{5mm}


\balance
\bibliographystyle{ACM-Reference-Format}
\bibliography{sample-base}

\appendix
\clearpage

\section*{Appendices} 

\subsection*{A. Experimental Setup}
In this section, we describe the detailed experimental setup, i.e., 
the pre-trained victim classifiers, the implementation of attack and defense methods that are evaluated in our work.

{\setlength{\parindent}{0cm}\textbf{Pre-trained Models:}} We use two two pre-trained models as the victim classifiers to conduct experiment.

\begin{itemize}
    \item ResNet-101. On ImageNet dataset, we use the pre-trained ResNet-101 network provided by TorchVision\footnote{https://github.com/pytorch/vision}, a python package consists of popular datasets, model architectures, and common image transformations for computer vision.
    
    \item ResNet-50. On CIFAR-10 dataset, we use the pre-trained ResNet-50 network\footnote{https://github.com/huyvnphan/PyTorch\_CIFAR10}, which is retrained on CIFAR-10 dataset based on modified TorchVision official implementation of popular CNN models.
\end{itemize}

{\setlength{\parindent}{0cm}\textbf{Attack Method Implementation:}}
For every attack methods, we use their implementations from open source repositories. It is noted that there are various toolbox to do  adversarial robustness experiment, and we choose the most widely used ones.

\begin{itemize}
    \item White-box attacks. Specifically, for FGSM \cite{FGSM}, PGD \cite{madry2018towards}, BIM \cite{Iteradv17}, MIFGSM \cite{MIFGSM} and C\&W \cite{carlini2017adversarial}attack methods, we use their implementation in Torchattacks \footnote{https://github.com/Harry24k/adversarial-attacks-pytorch}, a Pytorch repository for adversarial attacks. For DDN \cite{rony2019decoupling} attack, we adopt the implementation in foolbox\footnote{https://github.com/bethgelab/foolbox}, a Python toolbox to create adversarial examples.
    
    \item BPDA-based attack \cite{madry2018towards}. We use the BPDA-Wrapper in advTorch \footnote{https://github.com/BorealisAI/advertorch}, a toolbox for adversarial robustness research, to implement defense-aware white-box attack. For BPDA attack on baseline defense methods ComDefend \cite{jia2019comdefend}, NPR \cite{naseer2020self}, DIPDefend \cite{dai2020dipdefend} and ours, we adopt an identity mapping function to substitute the back-propagation computation, while a down-sampling operation is adopted on SR \cite{mustafa2019image} defense because the input and output image of SR defense are with different resolutions.
    
    \item Query-based attack. For HopSkipJump \cite{chen2020hopskipjumpattack} attack, we use the implementation from Adversarial Robustness Toolbox (ART) \footnote{https://github.com/Trusted-AI/adversarial-robustness-toolbox}, a Python library for Machine Learning Security.    
    
\end{itemize}

{\setlength{\parindent}{0cm}\textbf{Defense Method Implementation:}}
For the baseline defense methods ComDefend \footnote{https://github.com/jiaxiaojunQAQ/Comdefend} \cite{jia2019comdefend}, NPR\footnote{https://github.com/Muzammal-Naseer/NRP} \cite{naseer2020self} and SR\footnote{https://github.com/aamir-mustafa/super-resolution-adversarial-defense} \cite{mustafa2019image}, we use their implementation from GitHub websites given by the authors. For the DIPDefend  \cite{dai2020dipdefend} method, we use the implementation provided by authors.
In addition, we use Pytorch framework to implement the proposed defense method.

\subsection*{B. Network Architectures}
\begin{figure}[!htbp]
  \centering
  \includegraphics[width=0.95\linewidth]{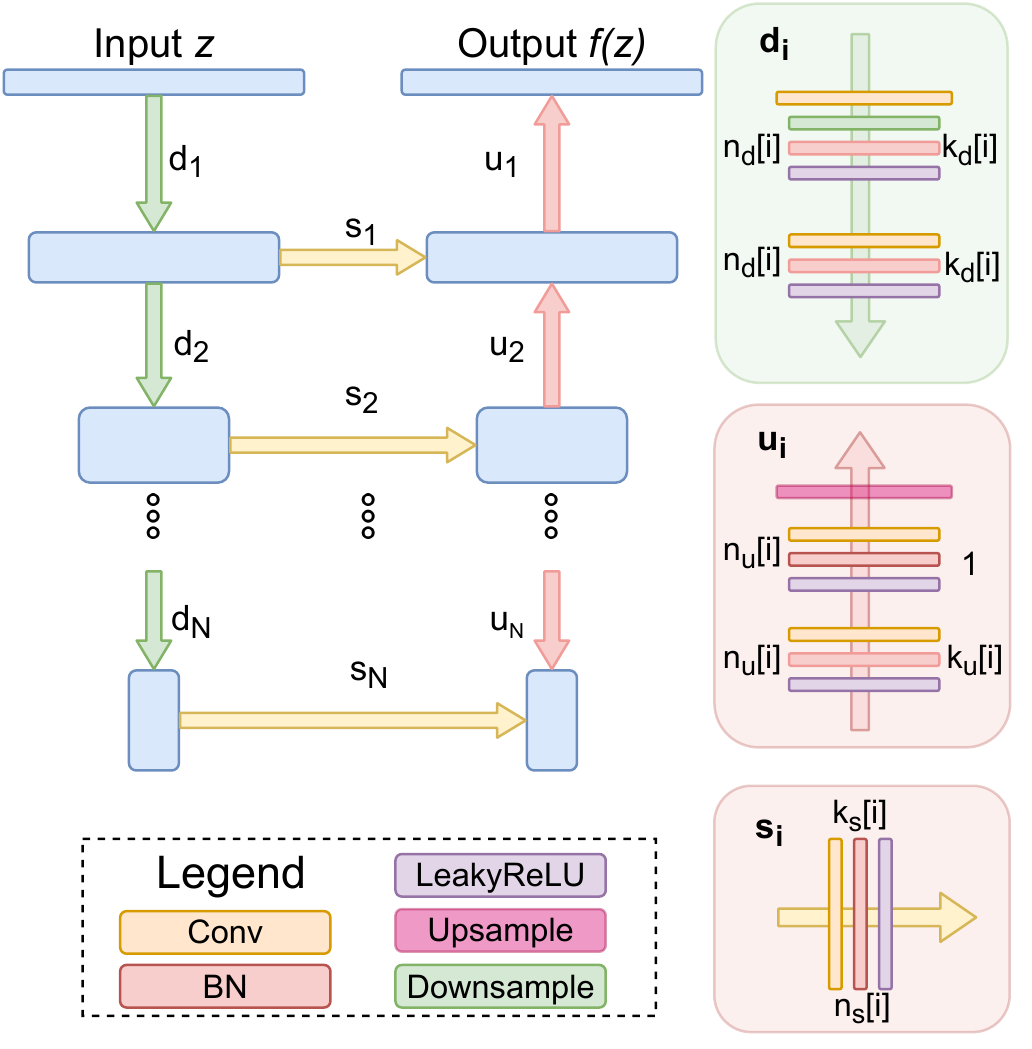}
  \caption{The architecture of DIP network used in the experiments. The U-Net architecture \cite{ronneberger2015u} with skip connections (yellow arrows) are used. $n_u, n_d, n_s$ correspond to the number of filters for the upsampling, downsampling and skip-connections, respectively. The values $k_u, k_d$ and $k_s$ correspond to the respective kernel sizes.} 
  \label{fig:dip_net_arch}
\end{figure}

Following the official DIP implementation in \cite{ulyanov2018deep}, we use the U-Net architecture with skip connection \cite{ronneberger2015u} as our DIP network. 
Fig. \ref{fig:dip_net_arch} shows the network structure in details. The hyper-parameters details of network architectures for experiments on CIFAR-10 and ImageNet are listed below using the notation introduced in Fig. \ref{fig:dip_net_arch}.
It is noted that these networks share the same structure except different hyper-parameters, i.e. number of filters and kernel sizes in every kind of layers, scale number for downsampling and upsampling. 
On the CIFAR-10 images, which resolution is $32\times32$, we found that a small network and a small iteration number $T$ is enough. While on ImageNet images, we adopt a larger network and iteration number $T$ to optimize the DIP network because the image size is $224\times224$.
Beside, we also describe the parameters of small/medium/large network in experiments on parameter sensitivity (Section \ref{sec:param_sen}).
In addition, the DIP model is initialized with random noise, and is trained with the Adam Optimizer. Learning rate is fixed to 0.01 and no learning rate decay is used.

\medskip\noindent\textbf{The DIP network architecture on CIFAR-10 dataset}
\\
\noindent\fbox{%
\parbox{0.97\linewidth}{%
$z \in \mathbb{R}^{4\times W \times H} \sim U(0,\frac{1}{30})$\\
$n_u = n_d = $ \texttt{[32, 32, 32]}\\
$k_u = k_d = $ \texttt{[3, 3, 3]}\\
$n_s = $ \texttt{[3, 3, 3]}\\
$k_s = $ \texttt{[1, 1, 1]}\\
\texttt{$T$ (Iteration Number)} $ = 1000$\\
\texttt{LR} $ = 0.01$\\
\texttt{Optimizer} $ = ADAM$\\
\texttt{upsampling} $ = $ \texttt{bilinear}
}}
\\

\medskip\noindent\textbf{The DIP network architecture on ImageNet dataset}
\\
\noindent\fbox{%
\parbox{0.97\linewidth}{%
$z \in \mathbb{R}^{4\times W \times H} \sim U(0,\frac{1}{100})$\\
$n_u = n_d = $ \texttt{[64, 64, 64, 64,64]}\\
$k_u = k_d = $ \texttt{[3, 3, 3, 3, 3]}\\
$n_s = $ \texttt{[4, 4, 4, 4, 4]}\\
$k_s = $ \texttt{[1, 1, 1, 1, 1]}\\
\texttt{$T$ (Iteration Number)} $ = 3000$\\
\texttt{LR} $ = 0.01$\\
\texttt{Optimizer} $ = ADAM$\\
\texttt{upsampling} $ = $ \texttt{bilinear}
}}
\\

\medskip\noindent\textbf{The \textit{small} DIP network architecture in \ref{sec:param_sen}}
\\
\noindent\fbox{%
\parbox{0.97\linewidth}{%
$z \in \mathbb{R}^{4\times W \times H} \sim U(0,\frac{1}{30})$\\
$n_u = n_d = $ \texttt{[16, 16]}\\
$k_u = k_d = $ \texttt{[2, 2]}\\
$n_s = $ \texttt{[4, 4]}\\
$k_s = $ \texttt{[1, 1, 1, 1]}\\
\texttt{$T$ (Iteration Number)} $ = 500/1000/1500$\\
\texttt{LR} $ = 0.01$\\
\texttt{Optimizer} $ = ADAM$\\
\texttt{upsampling} $ = $ \texttt{bilinear}
}}
\\

\medskip\noindent\textbf{The $'medium'$ DIP network architecture in \ref{sec:param_sen}}
\\
\noindent\fbox{%
\parbox{0.97\linewidth}{%
$z \in \mathbb{R}^{4\times W \times H} \sim U(0,\frac{1}{30})$\\
$n_u = n_d = $ \texttt{[32, 32, 32]}\\
$k_u = k_d = $ \texttt{[3, 3, 3]}\\
$n_s = $ \texttt{[3, 3, 3]}\\
$k_s = $ \texttt{[1, 1, 1]}\\
\texttt{$T$ (Iteration Number)} $ = 500/1000/1500$\\
\texttt{LR} $ = 0.01$\\
\texttt{Optimizer} $ = ADAM$\\
\texttt{upsampling} $ = $ \texttt{bilinear}
}}
\\

\medskip\noindent\textbf{The \textit{large} DIP network architecture in \ref{sec:param_sen}}
\\
\noindent\fbox{%
\parbox{0.97\linewidth}{%
$z \in \mathbb{R}^{4\times W \times H} \sim U(0,\frac{1}{30})$\\
$n_u = n_d = $ \texttt{[48, 48, 48, 48]}\\
$k_u = k_d = $ \texttt{[4, 4, 4, 4]}\\
$n_s = $ \texttt{[4, 4, 4, 4]}\\
$k_s = $ \texttt{[1, 1, 1, 1]}\\
\texttt{$T$ (Iteration Number)} $ = 500/1000/1500$\\
\texttt{LR} $ = 0.01$\\
\texttt{Optimizer} $ = ADAM$\\
\texttt{upsampling} $ = $ \texttt{bilinear}
}}

\subsection*{C. Visual Quality Comparison}
\begin{figure*}[!htbp]
  \centering
  \includegraphics[width=0.95\linewidth]{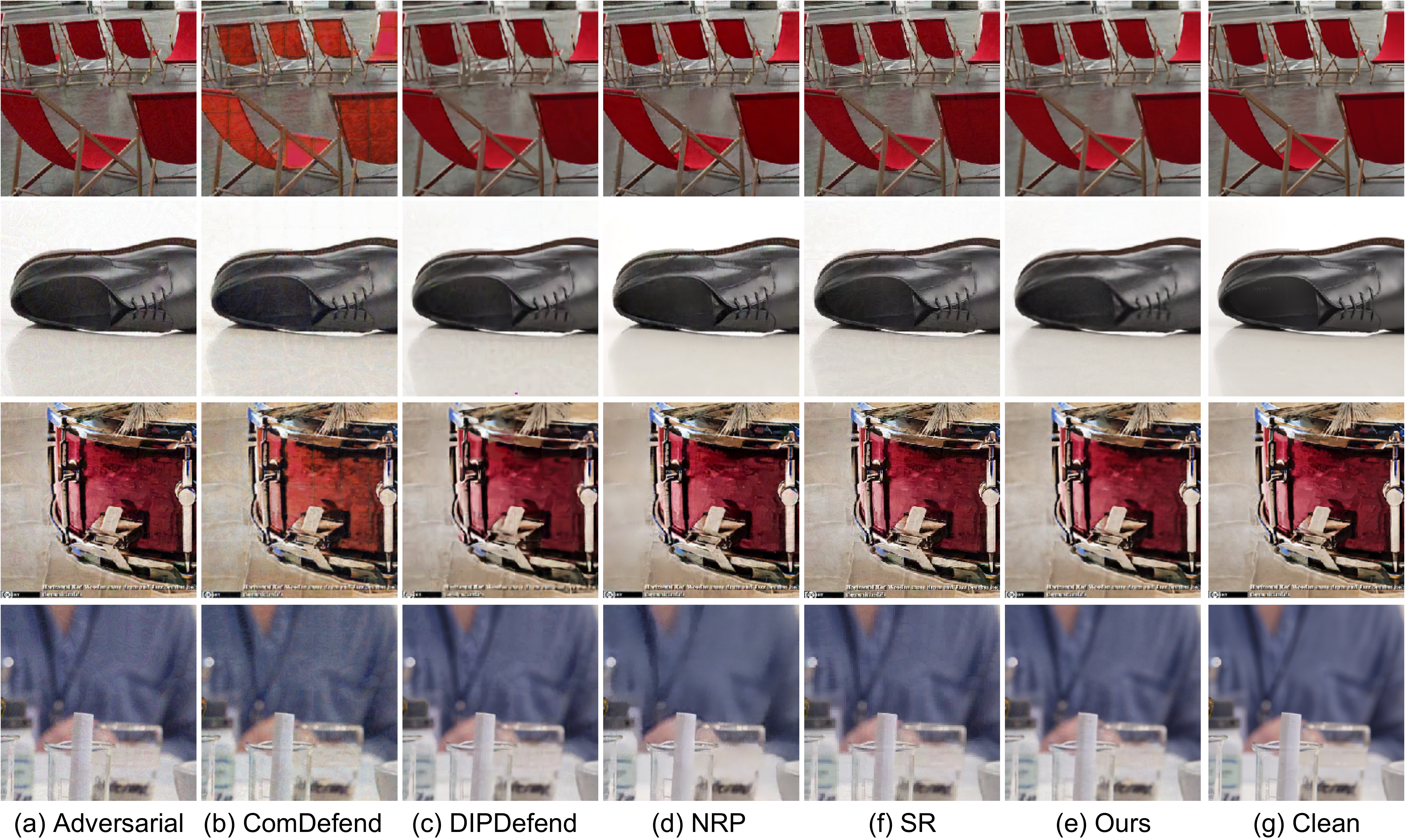}
  \caption{Visual quality comparison of image reconstruction results of different defense methods on ImageNet. Please zoom into digital images for better visual quality comparison. } 
  \label{fig:iqa_comp}
\end{figure*}

In this section, we visualize and compare the image quality of reconstructed images qualitatively. Specifically, the adversarial examples are generated by PGD attack with $\epsilon=8$. As shown in Fig. \ref{fig:iqa_comp}, the first column is adversarial image and the last column is the corresponding clean images. we can see that our reconstructed images can remove the adversarial noises in smooth area while remaining the structural details of clean images.

\subsection*{D. Visualizing the Defense Process}
\begin{figure*}[]
  \centering
  \includegraphics[width=0.9\linewidth]{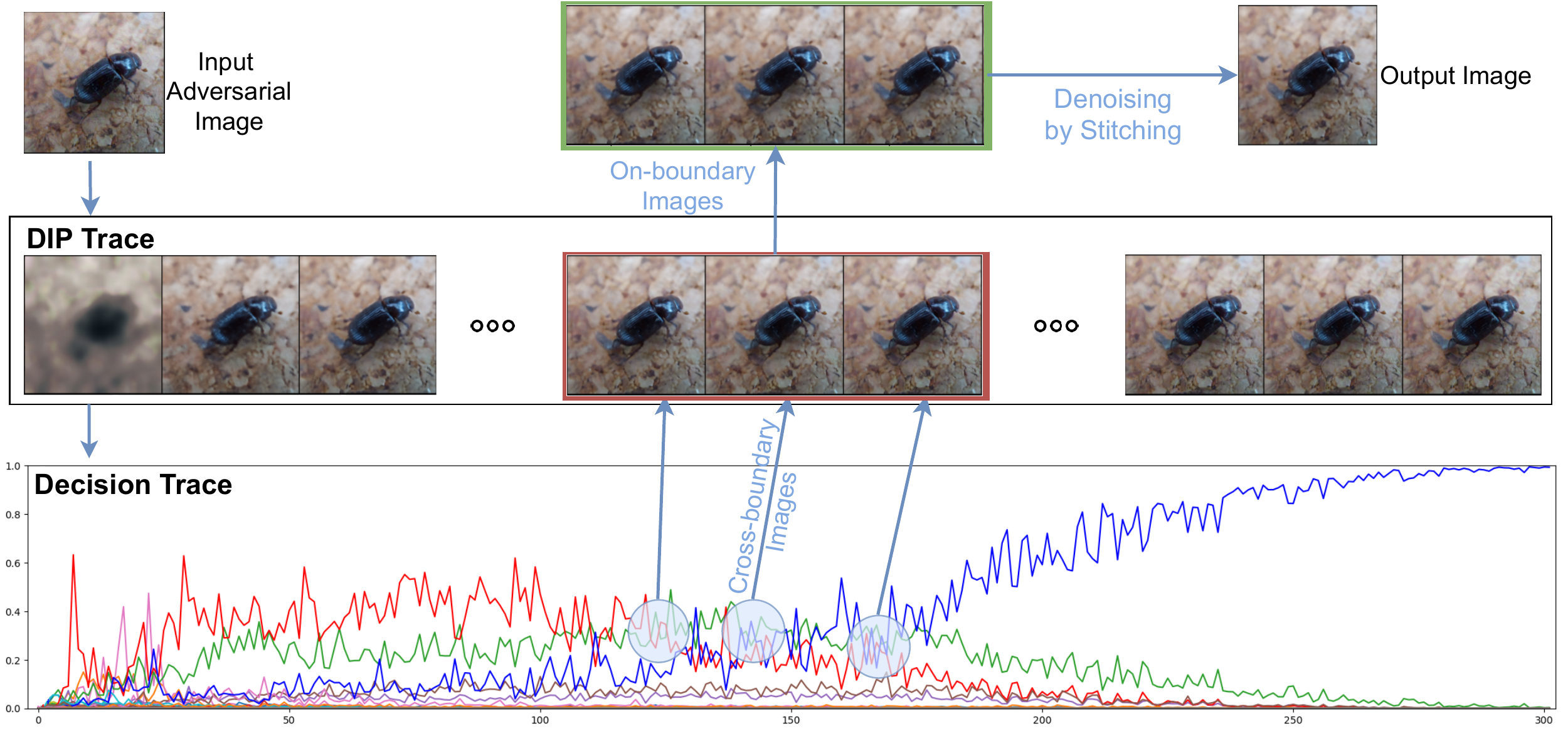}
  \caption{Visualization example of our defense method on an adversarial image from ImageNet. Please zoom into digital images for better observation of adversarial noise.} 
  \label{fig:vis_defense}
\end{figure*}

Our defense is essentially a denoising process in the decision space. To further understand our defense method intuitively, we visualize the process of our defense in Fig. \ref{fig:vis_defense}, showing an example of defense against an adversarial image.
First, we map the reconstructed image sequence of DIP network to the decision space, and then detect the cross-boundary images and on-boundary images. Finally, we reconstruct the output image by denoising these on-boundary images. 

As shown in \ref{fig:vis_defense}, the decision trace represents the victim classifier outputs as the number of DIP iterations increases. The red line denotes the class confidence of the original image, and the blue line is the adversarial class confidence. It can be seen from the figure that as the iteration increases, the output images of DIP will sequentially cross the boundary between the red and the green line, the boundary between the blue and the green line, and the boundary between the blue and the red line, and finally reach the class manifold where the adversarial image is located.
After detecting these cross-boundary images in the decision space, our defense method searches for images located on the boundary through linear interpolation. These on-boundary images are much likely to contain adversarial noise, so our final reconstructed image is synthesized by a two-stage denoising strategy on the on-boundary images.

\subsection*{E. Cross-model Transferability Defense}
This section aims to show the effectiveness of cross-model defense of our method. In this experiment, we reconstruct images based on the Inception-V3 model, then evaluate the defensive effectiveness on ResNet-50 (PGD on CIFAR-10). Compared with DIPDefend \cite{dai2020dipdefend}, the SOTA baseline, the proposed method still achieve better performance, i.e., ACC 0.81 vs. 0.76. This further validates the denoising effectiveness of our reconstruction method. 

\subsection*{F. Discussion on Overfitting Problems}
\cite{zhang2020geometry} tries to alleviate the overfitting issue during adversarial training. However, there are two major differences between \cite{zhang2020geometry} and our work. First, overfitting problems are different: work \cite{zhang2020geometry} addresses the overfitting problem in training DCNNs; while we resolve the overfitting issue in DIP-based defenses, an optimization process that requires no training. Correspondingly the strategies are different: work \cite{zhang2020geometry} utilizes different training instances during adversarial training; while we don’t rely on other different instances during defense.

Despite their differences, we find the adaptive re-weighting strategy in work \cite{zhang2020geometry} very interesting. After analysis, we believe the similar philosophy could be applied in our image stitching procedure. E.g., by assigning different weights for different on-manifold images, we may further enhance our defense method.

\end{document}